\documentclass[10pt,letterpaper]{article}

\usepackage{cogsci}

\cogscifinalcopy 

\usepackage{pslatex}
\usepackage{apacite}
\usepackage{float} 

\usepackage{booktabs}
\usepackage{url}
\usepackage{amsmath}
\usepackage{graphicx}
\usepackage{multirow}

\usepackage{xcolor}

\title{Is the Language Familiarity Effect gradual ? A computational modelling approach}

 \author{{\large \bf Maureen de Seyssel$^1$$^2$ (maureen.deseyssel@gmail.com)} \\
{\large \bf Guillaume Wisniewski$^2$ (guillaume.wisniewski@u-paris.fr)}\\
{\large \bf Emmanuel Dupoux$^1$ (emmanuel.dupoux@gmail.com)} 
\AND {$^1$ Cognitive Machine Learning (ENS–CNRS–EHESS–INRIA–PSL Research University)}\\
{$^2$ Université Paris Cité, CNRS, Laboratoire de Linguistique Formelle}
{Paris, France}}

\DeclareMathOperator{\test}{Ts}
\DeclareMathOperator{\train}{Tr}

\begin{document}

\maketitle


\begin{abstract}
According to the Language Familiarity Effect (LFE), people are better at discriminating between speakers of their native language. Although this cognitive effect was largely studied in the literature, experiments have only been conducted on a limited number of language pairs and their results only show the presence of the effect without yielding a gradual measure that may vary across language pairs.  In this work, we show that the computational model of  LFE introduced by \citeA{thorburn_quantitative_2019} can address these two limitations. In a first experiment, we attest to this model's capacity to obtain a gradual measure of the LFE by replicating behavioural findings on native and accented speech. In a second experiment, we evaluate LFE on a large number of language pairs, including many which have never been tested on humans. We show that the effect is replicated across a wide array of languages, providing further evidence of its universality. Building on the gradual measure of LFE, we also show that languages belonging to the same family yield smaller scores, supporting the idea of an effect of language distance on LFE. 
\end{abstract}

\section{Introduction}

The Language Familiarity Effect (LFE) is a cognitive effect observed
in language processing, according to which people are better at
discriminating speakers who speak in their native language, compared to speakers of another unfamiliar language \cite{goggin1991role, johnson_abstraction_2018}.
Two views are commonly proposed to explain the LFE \cite{perrachione_2018}. According to the Phonetic Familiarity hypothesis, the lack of familiarity with the foreign language's lower levels of linguistic characteristics (rhythm, phonetics, acoustics) is enough to explain the effect. For proponents of the Linguistic Processing hypothesis, on the other hand, the effect is in great part explained by the lack of understanding (due to knowledge of lexicon and syntax). However, even in this second view, the role of low-level linguistic features is accepted \cite{bregman2014gradient, perrachione2015effects}.


\subsubsection{Methodological issues}

Although numerous experimental studies run in humans (henceforth behavioural studies) found evidence of the effect, the lack of systematicity  
makes it hard to compare the results directly  \cite{levi_methodological_2019}. First, the evaluation tasks used to assess the presence of LFE differ from one study to the next, ranging from identification tasks (voice line-up) to discrimination tasks (AX task). Critically, a same language pair evaluated on different tasks can yield opposite results regarding the presence of LFE \cite{levi_methodological_2019}. Another source of variability comes from the initial testing conditions. Two setups are principally used, the ``1 Group 2 Languages" (or 1G2L) and the ``2 Groups 1 Language" (or 2G1L). In the first, most common condition, participants are all native speakers of the same language and are evaluated on their ability to discriminate between speakers in both their native language and a second unfamiliar language. In the 2G1L condition, two groups of participants, native speakers of languages A and B, are tested on only the same language A. 

One more issue raised from behavioural studies is the restricted number of language pairs tested. Although this effect has been found over multiple language pairs, leading to qualifying the effect of universal (see \citeA{levi_methodological_2019, perrachione_2018} for reviews), it turns out that only a small number of languages was tested. 
For instance, only a handful of studies test a language pair that does not contain English \cite{koster1997different, johnson2011infant, perea2014ability}. 
In order to get more robust evidence of the universality of the effect, a wider array of languages must be tested.

\medskip
 
\subsubsection{LFE as a gradual effect}

Because of how the LFE has been evaluated behaviourally, it has mainly been presented
as either present or absent. Very few attempts have been made at looking at the effect gradually: \citeA{perrachione_2018} computed effect sizes in LFE experiments, but they are hardly comparable due to differences in setup. Having a systematic gradual measure would allow deeper analyses of specific conditions. Hence, we could directly compare different language pairs or different atypical populations on the LFE.  
Some studies looked into the role of language distance in LFE. For example, \cite{levi_methodological_2019} showed, in an extensive literature review, that language pairs both from the same and different rhythmic classes could yield an LFE. However, 
this does not allow a gradual ranking of language pairs. A few studies directly tested multiple languages with the same population, permitting ranked comparisons, assuming that phonologically similar languages yield better performance in speaker identification. However, these studies never tested more than three languages at a time, and conflicting results were found. \citeA{koster1997different} and \citeA{zarate2015multiple} results confirmed this assumption (testing Chinese, English and Spanish on German adult listeners and English, German and Mandarin on English adult listeners, respectively). However, no difference between phonologically similar (English and Dutch) and dissimilar (English and Mandarin) were found in infants by \cite{johnson2011infant}, thus a need for further studies on the question. 
Hence, there is a need for a way to test and rank in a systematic manner a large number of language pairs, varying in language similarity.

 
Additionally, having a gradual measure of the LFE can help analyse finer granularity than that of language differences. An existing example is the case of accented speech. Indeed, some studies found that, for a language pair A-B, if the test stimuli in language A are spoken by native speakers of language B, and therefore accented in B, the LFE can be reduced \cite{goggin1991role}, and even totally cancelled \cite{goldstein1981recognition}. This suggests that the LFE can be modulated by how heavily accented the speech is and, to a further extent, that acoustically similar dialects should give rise to smaller LFE, corroborating the idea that language distance plays a role in this cognitive effect. These results also show  that the effect is gradual, emphasising the need for a gradual measure.


\subsubsection{I-vectors as a model of LFE}

Recently, \citeA{thorburn_quantitative_2019} were able to computationally model the LFE using i-vectors \cite{dehak2010cosine}, an unsupervised algorithm that allows to compute a representation of whole speech utterances. Computational modelling of LFE can help circumvent some of the methodological problems presented earlier, and we believe it can help  compute a systematic, gradual and comparable measure of the effect. 

I-vectors models, typically used for speaker-identification applications in speech processing, consist in training a Gaussian Mixture Model on speech features of the train sets utterances to define a new
representation of the acoustic space. Then, projecting the components of highest variability onto a lower-dimensional space, we can create a new representation of speech (the i-vector) for all utterances from the train set. By extension, we can predict representations for novel utterances based on this operation. Furthermore, because computed at the utterance level rather than at a finer frame level, we capture the acoustic information that is representative of the utterance as a whole, such as speaker or language information. 
The lack of time-dependencies in the representation means that only low-level features of linguistics (rhythm, some phonology) are captured. Because of that, and the fact that training such models only necessitates a small amount of input data, the approach has mainly been proposed in models of infants' speech perception. Still, we believe i-vectors can equally model some aspects of adult speech perception that do not require access to higher levels of linguistics.

I-vectors were first proposed in the context of speech perception by \citeA{carbajal2016language} as a model of language discrimination. Still modelling language discrimination processes, \citeA{deseyssel2020does} showed that they also capture speaker information, even without relying on any supervised components usually present in speech processing applications of i-vectors.
Because the LFE depends on both language and speaker information, the i-vector model has the necessary attributes to model it, and this is indeed what showed \citeA{thorburn_quantitative_2019}. In their paper, the authors focused on the English-Japanese pair. They
showed that the scores from a speaker discrimination task carried out on i-vector representations extracted on both languages were significantly better when the i-vectors were extracted using a model trained on the same language than on another unfamiliar language, effectively replicating the effect found in humans. 


\subsection{Contributions}

One underlying contribution of this paper is a replication of the computational approach of \citeA{thorburn_quantitative_2019} on new speech stimuli and language pairs. This reinforces the validity of the i-vector approach to model the LFE. Most importantly, and as the main contribution, we inquire about the capacity of the approach to yield a gradual, comparable measure.

In a first experiment, we look into \emph{reproducing the human findings} according to which accented speech minimises the LFE compared to native speech. Precisely, we replicated an experiment from \cite{wester2012talker} testing LFE on two language pairs that are always accented in one of the languages.  
This first experiment also allows us to directly \emph{compare} a close language pair to a distant one.
We expect three primary outcomes~: a replication of the LFE on the native condition; an LFE is smaller or non-existent in the accented condition compared to the native condition; an LFE is smaller in the close language distance pair than in the distant one. Such results would corroborate with the idea of a gradual effect of the LFE, which we could measure using the i-vector approach, allowing for systematic comparisons.

Findings from the first experiment then lead us to generalise the method to \emph{additional language pairs}. In the second experiment, we evaluate the LFE on 36 language pairs, with many that have never been tested in humans. We can then systematically (1) test and (2) compare the LFE on a large number of languages pairs in a way that would be impossible behaviourally.  We expect the LFE to replicate on most of these pairs and to find an effect of language distance.  

\section{General methods}\label{section : gen_methods}

The methods presented here are common to both experiments. We use as an example a setup in which we want to evaluate the LFE on a language pair A, B. For each language, we have a set of speech utterances, split between a train and a test set. The former is used to train the models, and the latter is the evaluation stimuli used to test the presence of LFE. 

\subsection{Training pipeline}

We first extract Mel Frequency Cepstral Coefficients (MFCCs) \cite{mermelstein1976distance} for all utterances (train and test), with 13 coefficients including energy, along with double delta coefficients. 
We also include pitch information through computation of the fundamental frequency, as it is thought to be relevant in language discrimination \cite{lin2005language}.


We then train two i-vector models using the MFCCs from the train sets, one on language A (model A) and one on language B (model B), following the approach first proposed in \citeA{dehak2010cosine}. The only difference with the original i-vector approach is that we do not carry out a Linear Discriminant Analysis (LDA), originally aiming at maximising the distance between speakers and/or language \cite{kanagasundaram2011vector,dehak2011language}. Indeed, as in previous studies using i-vectors as models of speech perception \cite{carbajal2016language,deseyssel2020does, thorburn_quantitative_2019}, we ensure that the pipeline is unsupervised and therefore better suited to cognitive models.
Finally, we extract i-vector representations from the two test sets on both models. That is, we extract i-vectors using model A on tests sets A and B, and similarly for model B. This leaves us with four sets of i-vectors: language A trained on A, language A trained on B, language B trained on A and language B trained on B.

The models are trained with 128 (2,048) Gaussians and i-vectors of dimension 150 (400) in Experiment 1 (Experiment 2). The difference in parameters between the two experiments is explained by the larger number of speakers in the training sets of Experiment 2. Feature extraction, models training and i-vectors extraction were conducted using the Kaldi toolkit \cite{povey2011kaldi}.




\subsection{Evaluation}\label{methods:evaluation}

Following \citeA{thorburn_quantitative_2019}, we first use a machine ABX task \cite{schatz2013evaluating} to evaluate the capacity of a model to discriminate speakers. 
In this setup, we create triplets of three utterances from the same language:  $a$, $b$ and $x$, with $a$ and $x$ being pronounced by the same speaker and $b$ by a different speaker. If the Euclidean distance between the representations (i.e.\ i-vectors) of utterances
$a$ and~$x$ is larger than the distance between the representation of $b$ and~$x$, we consider that the model did not manage to discriminate between the speakers, and we count an error for this specific triplet. The ABX error score is the error rate estimated over all possible triplets in the test set.

This framework can be extended to evaluate the LFE by comparing the capacity of a model to discriminate between speakers in a `familiar' condition, in which the representation is learnt and tested on the same language, to its capacity to discriminate between speakers in an `unfamiliar' condition, in which test utterances are in a different language than the ones used to train the model. 
More precisely, we define the LFE score as follows: for a language
pair~$(A, B)$, we compute the ABX error rates for all four conditions ($Ts$ stands for \textit{test} and $Tr$ for \textit{training}): $\test(A)_{\train(A)}$, $\test(A)_{\train(B)}$, $\test(B)_{\train(B)}$ and $Ts(B)_{Tr(A)}$,
where $Ts(A)_{Tr(B)}$ corresponds to the evaluation of the ABX error rate on the language $A$ when the representation has been trained on language $B$.  We then average the scores in the `familiar' condition ($\test(A)_{\train(A)}$ and $\test(B)_{\train(B)}$, the test and train sets being matched in language), and the scores in the `unfamiliar' condition ($\test(A)_{\train(B)}$ and $\test(B)_{\train(A)}$ in which train and test languages are different). The LFE score is defined as the relative percentage increase from the `familiar' to the `unfamiliar' condition:
\begin{equation}
\textrm{LFE} = \frac{S_\textrm{diff} - S_\textrm{same}}{S_\textrm{same}} \label{eq_lfe} 
    \end{equation}
where:
\begin{align} 
S_\textrm{same} &= \frac{\test(A)_{\train(A)} + \test(B)_{\train(B)}}{2} \\
S_\textrm{diff} &= \frac{\test(A)_{\train(B)} + \test(B)_{\train(A)}}{2}
\end{align}

We use a Two-Sample Fisher-Pitman Permutation Test with Monte-Carlo sampling to test whether this effect is significant. The score is significant if discrimination scores in the $S_\textrm{same}$ and $S_\textrm{same}$ groups are significantly different.
A positive significant LFE score reflects an effect of language familiarity, with a higher ABX error rate in the non-familiar condition than in the familiar condition. 

Because we are looking at the LFE on the language pair symmetrically, that is analogous to a `2 groups 2 languages' (or `2G2L') approach, (two groups of participants, native in two different languages, are tested on both languages). Hence, we are controlling for any biases due to a specific training set yielding better speaker discrimination performance, and thus singling out the actual LFE process.
This is a more robust evaluation setup than what is commonly done in behavioural work, where LFE is looked into from the perspective of a single language only.  

\section{Experiment 1: LFE and accented speech}\label{Section : exp1}

First, we focus on two language pairs, English-Finnish and English-German. For each of these pairs, we compare a “native” setup, where all tested speakers are native in the languages, and an “accented” setup, with English utterances being spoken by non-native speakers, hence Finnish accented or German-accented.

\subsection{Materials}

We retrieved audiobooks in English, German and Finnish from the LibriVox project\footnote{https://librivox.org} using the \texttt{Libri-Light} tools \cite{kahn2020libri}, and used a Voice Activity Detection model \cite{lavechin2020open} to segment speech.  
We then created for each language a 10 hours training set, balanced equally between 10 speakers. 

The test sets were built from the EMIME bilingual corpus \cite{wester2010emime}, which contains English, German and Finnish read speech uttered by native speakers, as well as English spoken by German and Finnish speakers, and is therefore accented. We built five different test sets: native Finnish, native German, native English, Finnish-accented English and German-accented English. Each test set is balanced equally between 12 speakers and has an average duration of 25 min (348 utterances). See Table \ref{Table: exp1 data} for more information.


\begin{table}
\begin{center} 
\caption{Summary of test sets in Experiment 1.}
\label{Table: exp1 data}
\vskip 0.12in
\begin{tabular}{@{}lllll@{}}
\toprule
Language & Accent type       & \begin{tabular}[c]{@{}l@{}}N speakers \\ (N male)\end{tabular} & \begin{tabular}[c]{@{}l@{}}Mean (SD)\\ utt dur (in s)\end{tabular}\\ \midrule
English  & native            & 12 (6)              & 3.21 \textit{(1.04)}        \\
         & Finnish  & 12 (6)             & 4.37 \textit{(1.48)}         \\
         & German   & 12 (6)            & 4.56 \textit{(1.52)}        \\
Finnish  & native            & 12 (6)              & 4.6 \textit{(1.29)}      \\
German   & native            & 12 (6)              & 4.6 \textit{(1.32)}     \\ \bottomrule
\end{tabular}
\end{center} 
\end{table}

\subsection{Results}

We calculated the LFE score on four language pairs following the procedure presented in the General Methods: native English and native German; native English and native Finnish; German-accented English and native German; Finnish-accented English and native Finnish. We refer to the two first pairs as \emph{native} and the two last as \emph{accented}.

\begin{table}[H]
\begin{center} 
\caption{LFE scores on the native and accented conditions for both language pairs in Experiment 1. 
Significance was estimated using a Two tailed Paired Fisher-Pitman Permutation Test with Monte-Carlo sampling (*: p$<$.05)
}
\label{Table: emime-lib_lfe-accented}
\vskip 0.12in
\begin{tabular}{lrr}
\toprule
\multirow{2}{*}{Language Pair} & \multicolumn{2}{c}{LFE (\%)}                              \\
                               & \multicolumn{1}{l}{native} & \multicolumn{1}{l}{accented} \\ \midrule
English - Finnish              &\textbf{+19.21*}                     & -8.62                        \\
English - German               &\textbf{+10.77 *}                    & -1.1                         \\\bottomrule
\end{tabular}
\end{center}
\end{table}
\vskip -0.in




The first thing to notice from Table \ref{Table: emime-lib_lfe-accented} is that both language pairs yield a \textit{significant} LFE score in the native condition (the familiar models yield better discrimination scores than the unfamiliar ones, p$<$.05 in both pairs), giving further support to the i-vector approach as a good model of LFE. 
Moreover, the LFE score is higher in the English-Finnish pair than in the English-German pair, suggesting that the distance between languages could modulate the LFE.



In the accented condition, there is no longer a significant difference between the familiar and unfamiliar models' scores on language discrimination, and this on both language pairs. Hence, whilst the LFE scores indicate the effect was present in the native conditions, it is no longer the case in the accented condition, that is, when one of the two languages is uttered with an accent from the other language. These results, which replicate the behavioural findings from \citeA{wester2012talker} as well as previous studies on accents, suggest that we can use the i-vector models to obtain a gradual measure of the LFE. 


\section{Experiment 2: Testing LFE on many language pairs}

Results from the first experiment not only validate further the i-vector approach as a good model of the LFE, but they also suggest that the resulting measure is gradual and thus comparable. Furthermore, they suggest that there might be an effect of language distance on LFE. 

In this second experiment, we generalise the experiment to  many language pairs, including pairs that have not been tested on humans. This allows us to 1) verify the universality of the effect, 2) make use of the gradual measure to compare pairs with varying language distances.

\subsection{Materials}

We used stimuli from the \texttt{CommonVoice 6.1} (CV) corpus \cite{ardila2019common}, which gathers read speech from a large number of languages. 
We selected nine languages (those for which we had enough data) and generated, for each of them, training sets of 15 hours split between 60 speakers and test sets of 30 minutes split between 20 speakers (see Table \ref{Table: Datasets CV} for the complete list). The high number of speakers is closer to the setup proposed by \citeA{thorburn_quantitative_2019} than in the first experiment and ensures more variability in the training set, leading to a more robust model. 

\begin{table}[h!]
\begin{center} 
\caption{Summary of languages in Experiment 2. Train sets have an average duration of 15 hours (60 speakers) and test sets have an average total duration of 30mn (20 speakers).}
\label{Table: Datasets CV}
\vskip 0.12in
\begin{tabular}{@{}lllll@{}}
\toprule
\textbf{Language} & \textbf{ISO} &\textbf{ Avg utt dur (s)} &\textbf{Family} \\ \midrule
Catalan        & cat &    5.10 \textit{(SD=1.82)}  & indo-european      \\
Welsh            & cy & 4.52 \textit{(SD=1.67)}& indo-european\\
German          & deu &  4.42 \textit{(SD=1.50)} & indo-european\\
English           & eng &  4.81  \textit{(SD=1.74}) & indo-european\\
Farsi          & fas  & 3.80 \textit{(SD=1.42)}  & indo-european \\
French           & fra &   4.78 \textit{(SD=1.51)}& indo-european\\
Italian        & ita   &   5.35 \textit{(SD=1.73)}  & indo-european\\
Kabyle           & kab & 3.38 \textit{(SD=1.23) }& afro-asiatic \\
Kinyarwanda           & kin    &   5.14\textit{(SD=1.80)} & niger-congo  \\\bottomrule
\end{tabular}
\end{center}
\end{table}

\subsection{Results}


Models were trained on the nine languages, and evaluation was run on all possible language pairs, yielding 36 LFE scores.

\par
\begin{figure}[hptp]
\begin{center}
\includegraphics[trim={0 0 0 3.05cm}, clip,  scale=0.46, keepaspectratio]{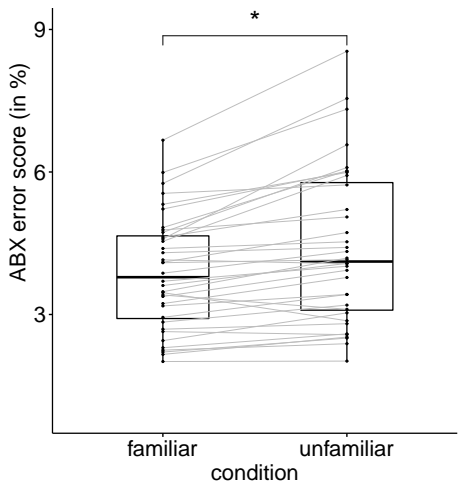}
\end{center}
  \caption{Speaker ABX error scores averaged across the 36 language pairs in the CV dataset. LFE score = 13.78. The asterisks on top illustrate the significance level (*, 95\% CI).} 
  \label{Figure: cv_all_paired}
\label{sample-figure}
\end{figure}

 Speaker ABX scores averaged across all pairs are presented in Figure \ref{Figure: cv_all_paired}, and detailed LFE scores are available in Table \ref{Table: LFE CV}.
Speaker discrimination scores are overall significantly higher in the `familiar' condition than in the `unfamiliar' one, with a mean LFE of  13.78 (significance was calculated using a 95\% confidence interval with bootstrapping on languages, with 10,000 permutations, CI = [2.71–18.55]). These results corroborate the idea of a universal LFE that can be expected on language pairs that were not tested on humans. 
However, the difference is not systematically significant in every pair, with one language pair (German-Welsh) yielding a significant inverse LFE. 



\par
\begin{figure}[hptp]
\begin{center}
\includegraphics[ scale=0.40, keepaspectratio]{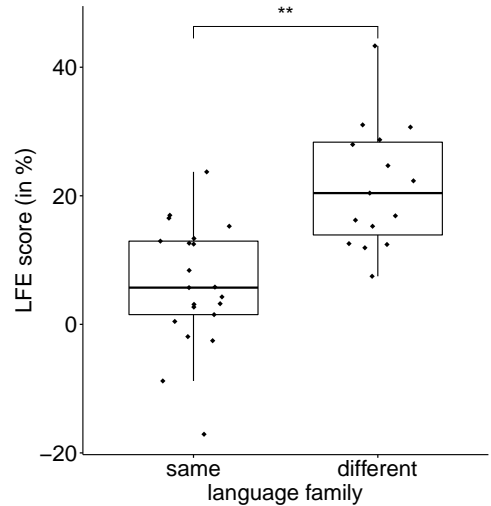}

\end{center}
  \caption{LFE scores averaged across the `same family' and `different family' conditions. The asterisks on top illustrate the significance level (**, 99\% CI).}
  \label{Figure: fam_abx}
\label{sample-figure}
\end{figure}

 We then divided the language pairs into two groups: the `same family' and the `different family', based on whether the languages in the pair belong or not to the same language family (as defined by the WALS typology \cite{wals}, see Table \ref{Table: Datasets CV}). As shown in Figure \ref{Figure: fam_abx}, the LFE scores from `same family' pairs (\textit{M}=21.46, \textit{SD}=9.62), \textit{N}=15 are significantly lower than the `different family' pairs (\textit{}M=6.13, \textit{SD}=9.47, \textit{N}=21) (significance was tested using a 99\% confidence interval with bootstrapping on language within family with 10,000 permutations, CI = [7.28,29.07]). 

\begin{table*}[htpb]
\begin{center} 
\caption{LFE scores for all possible CommonVoice language pairs. Two tailed Paired Fisher-Pitman Permutation Test with Monte-Carlo sampling with Bonferroni correction (*: p$<$.05; **: p$<$.005)}
\label{Table: LFE CV}
\vskip 0.12in
\begin{tabular}{lrrrrllll}
             & \multicolumn{1}{l}{\textbf{ca}} & \multicolumn{1}{l}{\textbf{cy}} & \multicolumn{1}{l}{\textbf{de}} & \multicolumn{1}{l}{\textbf{en}} & \textbf{fa}                          & \textbf{fr}                          & \textbf{it}                          & \textbf{kab}                         \\
\textbf{cy}  & 8.39                            & \multicolumn{1}{l}{}            & \multicolumn{1}{l}{}            & \multicolumn{1}{l}{}            &                                      &                                      &                                      &                                      \\
\textbf{de}  & 12.49              & \textit{-17.10**}                        & \multicolumn{1}{l}{}            & \multicolumn{1}{l}{}            &                                      &                                      &                                      &                                      \\
\textbf{en}  & 5.80                            & -8.81                        & 0.45                            & \multicolumn{1}{l}{}            &                                      &                                      &                                      &                                      \\
\textbf{fa}  & \textbf{16.53*}               & 1.51                            & 4.27                            & -2.55                           &                                      &                                      &                                      &                                      \\
\textbf{fr}  & \textbf{23.72**}                & 12.95              & \textbf{13.36**}               & \textbf{16.97*}                & \multicolumn{1}{r}{{\textbf{12.63**}}} &                                      &                                      &                                      \\
\textbf{it}  & 3.11                            & 3.21                            & -1.91                           & \textbf{15.27**}               & \multicolumn{1}{r}{5.71}             & \multicolumn{1}{r}{2.68}             &                                      &                                      \\
\textbf{kab} & \textbf{20.42**}                & 12.42                & \textbf{16.88**}                & 7.48                            & \multicolumn{1}{r}{\textbf{11.92**}} & \multicolumn{1}{r}{\textbf{16.21**}} & \multicolumn{1}{r}{\textbf{12.55**}} &                                      \\
\textbf{rw}  & \textbf{24.69**}                & \textbf{22.32**}                & \textbf{43.32**}               & \textbf{30.68**}                & \multicolumn{1}{r}{\textbf{15.26**}} & \multicolumn{1}{r}{\textbf{28.71**}} & \multicolumn{1}{r}{\textbf{27.97**}} & \multicolumn{1}{r}{\textbf{31.04**}}
\end{tabular}
\end{center}
\end{table*}

\section{Discussion}

In the first experiment, we successfully replicated results from \citeA{thorburn_quantitative_2019} by showing that the i-vector approach yields a significantly positive LFE score on two new language pairs (native condition). Moreover, we further validated this model by replicating another behaviour observed in humans, that is, the fact that the LFE can be diminished or cancelled with accented speech \cite{goldstein1981recognition, thompson1987language}. Specifically, the stimuli we use in our `accented' conditions come from the same dataset as in \cite{wester2012talker}, in which they also found no significant effect of LFE in humans, neither in the English-German nor in the English-Finnish pair. 


Differences in the LFE between accented and native speech support the idea of the LFE as a gradual effect that can be modulated by languages variations. Current psycholinguistic setups make such changes hard to capture in humans, but our results suggest that with the i-vector approach, we can capture and grade such changes. Being able to capture this granularity allows one to investigate the role of different variables on the LFE, with the most obvious one being that of language distance. As discussed earlier, the role of language distance  
is hard to analyse in behavioural experiments. However, results from Experiment 1 do suggest an effect, with the close distance language pair (English-German) yielding a lower LFE than the distant language pair (English-Finnish).

In the second experiment, we tested the model on a larger number of language pairs, of which many have never been tested on humans. We could compute comparable LFE scores for each of the 36 language pairs. We note that the LFE was present overall (across all pairs together), validating the approach again. However, not all language pairs yielded a significant effect. Multiple things could cause this: these specific pairs might not actually yield an LFE in humans, or the LFE might be too small in humans,
and the model is not sensitive enough to capture it. Regardless, we should replicate the experiment behaviourally, especially if the pair had not been tested on humans before. Finally, there might also be specific biases in the stimuli resulting in an absence of LFE. 
For example, two pairs, Welsh-English and Welsh-German, actually yielded a 
negative LFE score: the unfamiliar condition yielded better discrimination scores than the familiar one. However, it is likely that a large part of the Welsh utterances in the CV corpus was pronounced by English native speakers, which, in light of the previous results on accented speech and LFE, could partially explain the lack of LFE.  

Interesting results arose when the language pairs were divided into two groups: those in which both languages of the pair belong to the same family and those in which they do not. There was a significant difference in LFE scores between the two groups, with the same family language pairs yielding much lower LFE scores than the different family pairs. This corroborates with results from Experiment 1, suggesting that closer languages lead to a smaller LFE. 
The possibility for the LFE to be affected by language distance raises many interesting points regarding the cognitive processes behind this phenomenon. Commonalities and differences between languages can occur at various linguistic levels, but the i-vector approach only focuses on low-level cues (mainly phonology, prosody and phonotactics).  
Yet, it suggests an effect of language distance, supporting further the idea that the LFE is largely due to the familiarity with the phonetics and phonology of the foreign language, as proposed by the \emph{phonetic familiarity} hypothesis \cite{fleming_language-familiarity_2014, orena2015language}. Still, the distance between two languages at higher linguistic levels may also enhance the phenomenon.

To conclude, although it is not guaranteed that the language distance effect suggested by the model is equally present in humans, our results give us strong incentives to investigate this. This should be done in a systematic setup allowing for direct comparison of language pairs, potentially by designing a wide-scale online speaker discrimination study in many languages. Still, it is yet unclear whether we can obtain a fine enough gradual measure in humans.

\subsubsection{LFE score stability}

One of the central issues in computational modelling is the impact of data on the models. 
Here, we consider that the i-vector-based LFE score is stable in that it is not affected by changes in train or test sets.  
This is why we can confidently compare two conditions (languages, setup, number of speakers, recording condition) as long as all other factors are controlled for. However, we have not tested whether the results are prone to variations based on the train and evaluation stimuli, for example by running the same experiments on a new train or test sampled from the same original dataset. Only if the results are stable can we fully validate the approach. This stability aspect also raises the question of how representative of a language the training sets are. Indeed, while humans have had years of being exposed to their native language, which allows them to build an internal language prototype, we only train the models on a few hours of data in the current approach. Despite being a considerable advantage in data collection, it also increases the probability for the model's prototype to be biased. In the second experiment, we purposely used a high number of speakers and diversity in the recording setup, and we recommend any further work to follow this lead. 


Finally, we would like to address the fact that the i-vector model was initially proposed as a model of infant perception \cite{carbajal2016language, deseyssel2020does}, and used as such in the scope of the LFE to support the evidence that the effect only requires low-level linguistic knowledge and is present in infants \cite{thorburn_quantitative_2019}. Here, we focused on the LFE in general, without restriction to a specific age group. Indeed, although the present model only requires knowledge of the acoustics of the language, it still reproduces behavioural results found in adults, and we can only assume that adding higher up knowledge that makes use of language's understanding 
will only reinforce the effect found here. Therefore, even if the model could be completed by adding such features, the present approach can still be seen as a model of LFE in both infants and adults and has the advantage of only requiring very little data. 

\section{Conclusion}

To conclude, our results further validate the i-vector approach as a good model of the LFE by replicating \citeA{thorburn_quantitative_2019} on novel languages and replicating human experiments on accented speech. These results on accents also suggest that the effect can be modulated, hence gradual. The i-vector model allows computation of gradual LFE scores, meaning that we can then directly compare different conditions.
We showed further evidence of the universality of the effect by evaluating it on a large number of pairs systematically, in a way that can only be done computationally. We also found an effect of language distance, with larger LFE yielded when the two languages are dissimilar. These results should be replicated with humans in a setup allowing systematic evaluation, and attention should be given to the design of such an experiment. 
Finally, a more thorough analysis of the stability of the model depending on the training set could be done to ensure that the scores are fully stable and that different data would not give different scores, skewing the comparisons.

\section{Acknowledgements}
MS's work was partly funded by l'Agence de l'Innovation de Défense and performed using HPC resources from GENCI-IDRIS (Grant 20XX-AD011012315). ED in his EHESS role was supported in part by the Agence Nationale pour la Recherche (ANR-17-EURE-0017 Frontcog, ANR-10-IDEX-0001-02 PSL*, ANR-19-P3IA-0001 PRAIRIE 3IA Institute) and a grant from CIFAR (Learning in Machines and Brains).

\bibliographystyle{apacite}

\setlength{\bibleftmargin}{.125in}
\setlength{\bibindent}{-\bibleftmargin}

\bibliography{CogSci_bib}

\begin{thebibliography}{}

\bibitem [\protect \citeauthoryear {%
Ardila%
\ \protect \BOthers {.}}{%
Ardila%
\ \protect \BOthers {.}}{%
{\protect \APACyear {2019}}%
}]{%
ardila2019common}
\APACinsertmetastar {%
ardila2019common}%
\begin{APACrefauthors}%
Ardila, R.%
, Branson, M.%
, Davis, K.%
, Henretty, M.%
, Kohler, M.%
, Meyer, J.%
\BDBL {}Weber, G.%
\end{APACrefauthors}%
\unskip\
\newblock
\APACrefYearMonthDay{2019}{}{}.
\newblock
{\BBOQ}\APACrefatitle {Common voice: A massively-multilingual speech corpus}
  {Common voice: A massively-multilingual speech corpus}.{\BBCQ}
\newblock
\APACjournalVolNumPages{arXiv preprint arXiv:1912.06670}{}{}{}.
\PrintBackRefs{\CurrentBib}

\bibitem [\protect \citeauthoryear {%
Bregman%
\ \BBA {} Creel%
}{%
Bregman%
\ \BBA {} Creel%
}{%
{\protect \APACyear {2014}}%
}]{%
bregman2014gradient}
\APACinsertmetastar {%
bregman2014gradient}%
\begin{APACrefauthors}%
Bregman, M\BPBI R.%
\BCBT {}\ \BBA {} Creel, S\BPBI C.%
\end{APACrefauthors}%
\unskip\
\newblock
\APACrefYearMonthDay{2014}{}{}.
\newblock
{\BBOQ}\APACrefatitle {Gradient language dominance affects talker learning}
  {Gradient language dominance affects talker learning}.{\BBCQ}
\newblock
\APACjournalVolNumPages{Cognition}{130}{1}{85--95}.
\PrintBackRefs{\CurrentBib}

\bibitem [\protect \citeauthoryear {%
Carbajal%
, Dawud%
, Thiolli{\`e}re%
\BCBL {}\ \BBA {} Dupoux%
}{%
Carbajal%
\ \protect \BOthers {.}}{%
{\protect \APACyear {2016}}%
}]{%
carbajal2016language}
\APACinsertmetastar {%
carbajal2016language}%
\begin{APACrefauthors}%
Carbajal, M\BPBI J.%
, Dawud, A.%
, Thiolli{\`e}re, R.%
\BCBL {}\ \BBA {} Dupoux, E.%
\end{APACrefauthors}%
\unskip\
\newblock
\APACrefYearMonthDay{2016}{}{}.
\newblock
{\BBOQ}\APACrefatitle {The “language filter” hypothesis: A feasibility
  study of language separation in infancy using unsupervised clustering of
  i-vectors} {The “language filter” hypothesis: A feasibility study of
  language separation in infancy using unsupervised clustering of
  i-vectors}.{\BBCQ}
\newblock
\BIn{} \APACrefbtitle {2016 joint ieee international conference on development
  and learning and epigenetic robotics (icdl-epirob)} {2016 joint ieee
  international conference on development and learning and epigenetic robotics
  (icdl-epirob)}\ (\BPGS\ 195--201).
\PrintBackRefs{\CurrentBib}

\bibitem [\protect \citeauthoryear {%
Dehak%
\ \protect \BOthers {.}}{%
Dehak%
\ \protect \BOthers {.}}{%
{\protect \APACyear {2010}}%
}]{%
dehak2010cosine}
\APACinsertmetastar {%
dehak2010cosine}%
\begin{APACrefauthors}%
Dehak, N.%
, Dehak, R.%
, Glass, J\BPBI R.%
, Reynolds, D\BPBI A.%
, Kenny, P.%
\BCBL {}\ \BOthersPeriod {.}\end{APACrefauthors}%
\unskip\
\newblock
\APACrefYearMonthDay{2010}{}{}.
\newblock
{\BBOQ}\APACrefatitle {Cosine similarity scoring without score normalization
  techniques.} {Cosine similarity scoring without score normalization
  techniques.}{\BBCQ}
\newblock
\BIn{} \APACrefbtitle {Odyssey} {Odyssey}\ (\BPG~15).
\PrintBackRefs{\CurrentBib}

\bibitem [\protect \citeauthoryear {%
Dehak%
, Torres-Carrasquillo%
, Reynolds%
\BCBL {}\ \BBA {} Dehak%
}{%
Dehak%
\ \protect \BOthers {.}}{%
{\protect \APACyear {2011}}%
}]{%
dehak2011language}
\APACinsertmetastar {%
dehak2011language}%
\begin{APACrefauthors}%
Dehak, N.%
, Torres-Carrasquillo, P\BPBI A.%
, Reynolds, D.%
\BCBL {}\ \BBA {} Dehak, R.%
\end{APACrefauthors}%
\unskip\
\newblock
\APACrefYearMonthDay{2011}{}{}.
\newblock
{\BBOQ}\APACrefatitle {Language recognition via i-vectors and dimensionality
  reduction} {Language recognition via i-vectors and dimensionality
  reduction}.{\BBCQ}
\newblock
\BIn{} \APACrefbtitle {Twelfth annual conference of the international speech
  communication association.} {Twelfth annual conference of the international
  speech communication association.}
\PrintBackRefs{\CurrentBib}

\bibitem [\protect \citeauthoryear {%
de Seyssel%
\ \BBA {} Dupoux%
}{%
de Seyssel%
\ \BBA {} Dupoux%
}{%
{\protect \APACyear {2020}}%
}]{%
deseyssel2020does}
\APACinsertmetastar {%
deseyssel2020does}%
\begin{APACrefauthors}%
de Seyssel, M.%
\BCBT {}\ \BBA {} Dupoux, E.%
\end{APACrefauthors}%
\unskip\
\newblock
\APACrefYearMonthDay{2020}{}{}.
\newblock
{\BBOQ}\APACrefatitle {Does bilingual input hurt? A simulation of language
  discrimination and clustering using i-vectors} {Does bilingual input hurt? a
  simulation of language discrimination and clustering using i-vectors}.{\BBCQ}
\newblock
\BIn{} \APACrefbtitle {CogSci 2020-42nd Annual Virtual Meeting of the Cognitive
  Science Society.} {Cogsci 2020-42nd annual virtual meeting of the cognitive
  science society.}
\PrintBackRefs{\CurrentBib}

\bibitem [\protect \citeauthoryear {%
Dryer%
\ \BBA {} Haspelmath%
}{%
Dryer%
\ \BBA {} Haspelmath%
}{%
{\protect \APACyear {2013}}%
}]{%
wals}
\APACinsertmetastar {%
wals}%
\begin{APACrefauthors}%
Dryer, M\BPBI S.%
\BCBT {}\ \BBA {} Haspelmath, M.%
\end{APACrefauthors}%
\ (\BEDS).
\unskip\
\newblock
\APACrefYear{2013}.
\newblock
\APACrefbtitle {WALS Online} {Wals online}.
\newblock
\APACaddressPublisher{Leipzig}{Max Planck Institute for Evolutionary
  Anthropology}.
\newblock
\begin{APACrefURL} \url{https://wals.info/} \end{APACrefURL}
\PrintBackRefs{\CurrentBib}

\bibitem [\protect \citeauthoryear {%
Fleming%
, Giordano%
, Caldara%
\BCBL {}\ \BBA {} Belin%
}{%
Fleming%
\ \protect \BOthers {.}}{%
{\protect \APACyear {2014}}%
}]{%
fleming_language-familiarity_2014}
\APACinsertmetastar {%
fleming_language-familiarity_2014}%
\begin{APACrefauthors}%
Fleming, D.%
, Giordano, B\BPBI L.%
, Caldara, R.%
\BCBL {}\ \BBA {} Belin, P.%
\end{APACrefauthors}%
\unskip\
\newblock
\APACrefYearMonthDay{2014}{{\APACmonth{09}}}{}.
\newblock
{\BBOQ}\APACrefatitle {A language-familiarity effect for speaker discrimination
  without comprehension} {A language-familiarity effect for speaker
  discrimination without comprehension}.{\BBCQ}
\newblock
\APACjournalVolNumPages{Proceedings of the National Academy of
  Sciences}{111}{38}{13795--13798}.
\newblock
\begin{APACrefURL}
  [{2021-01-29}]\url{https://www.pnas.org/content/111/38/13795}
  \end{APACrefURL}
\newblock
\APACrefnote{Publisher: National Academy of Sciences Section: Social Sciences}
\newblock
\begin{APACrefDOI} \doi{10.1073/pnas.1401383111} \end{APACrefDOI}
\PrintBackRefs{\CurrentBib}

\bibitem [\protect \citeauthoryear {%
Goggin%
, Thompson%
, Strube%
\BCBL {}\ \BBA {} Simental%
}{%
Goggin%
\ \protect \BOthers {.}}{%
{\protect \APACyear {1991}}%
}]{%
goggin1991role}
\APACinsertmetastar {%
goggin1991role}%
\begin{APACrefauthors}%
Goggin, J\BPBI P.%
, Thompson, C\BPBI P.%
, Strube, G.%
\BCBL {}\ \BBA {} Simental, L\BPBI R.%
\end{APACrefauthors}%
\unskip\
\newblock
\APACrefYearMonthDay{1991}{}{}.
\newblock
{\BBOQ}\APACrefatitle {The role of language familiarity in voice
  identification} {The role of language familiarity in voice
  identification}.{\BBCQ}
\newblock
\APACjournalVolNumPages{Memory \& cognition}{19}{5}{448--458}.
\PrintBackRefs{\CurrentBib}

\bibitem [\protect \citeauthoryear {%
Goldstein%
, Knight%
, Bailis%
\BCBL {}\ \BBA {} Conover%
}{%
Goldstein%
\ \protect \BOthers {.}}{%
{\protect \APACyear {1981}}%
}]{%
goldstein1981recognition}
\APACinsertmetastar {%
goldstein1981recognition}%
\begin{APACrefauthors}%
Goldstein, A\BPBI G.%
, Knight, P.%
, Bailis, K.%
\BCBL {}\ \BBA {} Conover, J.%
\end{APACrefauthors}%
\unskip\
\newblock
\APACrefYearMonthDay{1981}{}{}.
\newblock
{\BBOQ}\APACrefatitle {Recognition memory for accented and unaccented voices}
  {Recognition memory for accented and unaccented voices}.{\BBCQ}
\newblock
\APACjournalVolNumPages{Bulletin of the Psychonomic Society}{17}{5}{217--220}.
\PrintBackRefs{\CurrentBib}

\bibitem [\protect \citeauthoryear {%
Johnson%
, Bruggeman%
\BCBL {}\ \BBA {} Cutler%
}{%
Johnson%
\ \protect \BOthers {.}}{%
{\protect \APACyear {2018}}%
}]{%
johnson_abstraction_2018}
\APACinsertmetastar {%
johnson_abstraction_2018}%
\begin{APACrefauthors}%
Johnson, E\BPBI K.%
, Bruggeman, L.%
\BCBL {}\ \BBA {} Cutler, A.%
\end{APACrefauthors}%
\unskip\
\newblock
\APACrefYearMonthDay{2018}{}{}.
\newblock
{\BBOQ}\APACrefatitle {Abstraction and the ({Misnamed}) {Language}
  {Familiarity} {Effect}} {Abstraction and the ({Misnamed}) {Language}
  {Familiarity} {Effect}}.{\BBCQ}
\newblock
\APACjournalVolNumPages{Cognitive Science}{42}{2}{633--645}.
\PrintBackRefs{\CurrentBib}

\bibitem [\protect \citeauthoryear {%
Johnson%
, Westrek%
, Nazzi%
\BCBL {}\ \BBA {} Cutler%
}{%
Johnson%
\ \protect \BOthers {.}}{%
{\protect \APACyear {2011}}%
}]{%
johnson2011infant}
\APACinsertmetastar {%
johnson2011infant}%
\begin{APACrefauthors}%
Johnson, E\BPBI K.%
, Westrek, E.%
, Nazzi, T.%
\BCBL {}\ \BBA {} Cutler, A.%
\end{APACrefauthors}%
\unskip\
\newblock
\APACrefYearMonthDay{2011}{}{}.
\newblock
{\BBOQ}\APACrefatitle {Infant ability to tell voices apart rests on language
  experience} {Infant ability to tell voices apart rests on language
  experience}.{\BBCQ}
\newblock
\APACjournalVolNumPages{Developmental Science}{14}{5}{1002--1011}.
\PrintBackRefs{\CurrentBib}

\bibitem [\protect \citeauthoryear {%
Kahn%
\ \protect \BOthers {.}}{%
Kahn%
\ \protect \BOthers {.}}{%
{\protect \APACyear {2020}}%
}]{%
kahn2020libri}
\APACinsertmetastar {%
kahn2020libri}%
\begin{APACrefauthors}%
Kahn, J.%
, Rivi{\`e}re, M.%
, Zheng, W.%
, Kharitonov, E.%
, Xu, Q.%
, Mazar{\'e}, P\BHBI E.%
\BDBL {}others%
\end{APACrefauthors}%
\unskip\
\newblock
\APACrefYearMonthDay{2020}{}{}.
\newblock
{\BBOQ}\APACrefatitle {Libri-light: A benchmark for asr with limited or no
  supervision} {Libri-light: A benchmark for asr with limited or no
  supervision}.{\BBCQ}
\newblock
\BIn{} \APACrefbtitle {ICASSP 2020-2020 IEEE International Conference on
  Acoustics, Speech and Signal Processing (ICASSP)} {Icassp 2020-2020 ieee
  international conference on acoustics, speech and signal processing
  (icassp)}\ (\BPGS\ 7669--7673).
\PrintBackRefs{\CurrentBib}

\bibitem [\protect \citeauthoryear {%
Kanagasundaram%
, Vogt%
, Dean%
, Sridharan%
\BCBL {}\ \BBA {} Mason%
}{%
Kanagasundaram%
\ \protect \BOthers {.}}{%
{\protect \APACyear {2011}}%
}]{%
kanagasundaram2011vector}
\APACinsertmetastar {%
kanagasundaram2011vector}%
\begin{APACrefauthors}%
Kanagasundaram, A.%
, Vogt, R.%
, Dean, D.%
, Sridharan, S.%
\BCBL {}\ \BBA {} Mason, M.%
\end{APACrefauthors}%
\unskip\
\newblock
\APACrefYearMonthDay{2011}{}{}.
\newblock
{\BBOQ}\APACrefatitle {I-vector based speaker recognition on short utterances}
  {I-vector based speaker recognition on short utterances}.{\BBCQ}
\newblock
\BIn{} \APACrefbtitle {Proceedings of the 12th Annual Conference of the
  International Speech Communication Association} {Proceedings of the 12th
  annual conference of the international speech communication association}\
  (\BPGS\ 2341--2344).
\PrintBackRefs{\CurrentBib}

\bibitem [\protect \citeauthoryear {%
K{\"o}ster%
, Schiller%
\BCBL {}\ \protect \BOthers {.}}{%
K{\"o}ster%
\ \protect \BOthers {.}}{%
{\protect \APACyear {1997}}%
}]{%
koster1997different}
\APACinsertmetastar {%
koster1997different}%
\begin{APACrefauthors}%
K{\"o}ster, O.%
, Schiller, N\BPBI O.%
\BCBL {}\ \BOthersPeriod {.}\end{APACrefauthors}%
\unskip\
\newblock
\APACrefYearMonthDay{1997}{}{}.
\newblock
{\BBOQ}\APACrefatitle {Different influences of the native language of a
  listener on speaker recognition} {Different influences of the native language
  of a listener on speaker recognition}.{\BBCQ}
\newblock
\APACjournalVolNumPages{Forensic Linguistics}{4}{}{18--28}.
\PrintBackRefs{\CurrentBib}

\bibitem [\protect \citeauthoryear {%
Lavechin%
, Bousbib%
, Bredin%
, Dupoux%
\BCBL {}\ \BBA {} Cristia%
}{%
Lavechin%
\ \protect \BOthers {.}}{%
{\protect \APACyear {2020}}%
}]{%
lavechin2020open}
\APACinsertmetastar {%
lavechin2020open}%
\begin{APACrefauthors}%
Lavechin, M.%
, Bousbib, R.%
, Bredin, H.%
, Dupoux, E.%
\BCBL {}\ \BBA {} Cristia, A.%
\end{APACrefauthors}%
\unskip\
\newblock
\APACrefYearMonthDay{2020}{}{}.
\newblock
{\BBOQ}\APACrefatitle {An open-source voice type classifier for child-centered
  daylong recordings} {An open-source voice type classifier for child-centered
  daylong recordings}.{\BBCQ}
\newblock
\APACjournalVolNumPages{arXiv preprint arXiv:2005.12656}{}{}{}.
\PrintBackRefs{\CurrentBib}

\bibitem [\protect \citeauthoryear {%
Levi%
}{%
Levi%
}{%
{\protect \APACyear {2019}}%
}]{%
levi_methodological_2019}
\APACinsertmetastar {%
levi_methodological_2019}%
\begin{APACrefauthors}%
Levi, S\BPBI V.%
\end{APACrefauthors}%
\unskip\
\newblock
\APACrefYearMonthDay{2019}{}{}.
\newblock
{\BBOQ}\APACrefatitle {Methodological considerations for interpreting the
  language familiarity effect in talker processing} {Methodological
  considerations for interpreting the language familiarity effect in talker
  processing}.{\BBCQ}
\newblock
\APACjournalVolNumPages{Wiley Interdisciplinary Reviews: Cognitive
  Science}{10}{2}{e1483}.
\PrintBackRefs{\CurrentBib}

\bibitem [\protect \citeauthoryear {%
Lin%
\ \BBA {} Wang%
}{%
Lin%
\ \BBA {} Wang%
}{%
{\protect \APACyear {2005}}%
}]{%
lin2005language}
\APACinsertmetastar {%
lin2005language}%
\begin{APACrefauthors}%
Lin, C\BHBI Y.%
\BCBT {}\ \BBA {} Wang, H\BHBI C.%
\end{APACrefauthors}%
\unskip\
\newblock
\APACrefYearMonthDay{2005}{}{}.
\newblock
{\BBOQ}\APACrefatitle {Language identification using pitch contour information}
  {Language identification using pitch contour information}.{\BBCQ}
\newblock
\BIn{} \APACrefbtitle {Proceedings.(ICASSP'05). IEEE International Conference
  on Acoustics, Speech, and Signal Processing, 2005.} {Proceedings.(icassp'05).
  ieee international conference on acoustics, speech, and signal processing,
  2005.}\ (\BVOL~1, \BPGS\ I--601).
\PrintBackRefs{\CurrentBib}

\bibitem [\protect \citeauthoryear {%
Mermelstein%
}{%
Mermelstein%
}{%
{\protect \APACyear {1976}}%
}]{%
mermelstein1976distance}
\APACinsertmetastar {%
mermelstein1976distance}%
\begin{APACrefauthors}%
Mermelstein, P.%
\end{APACrefauthors}%
\unskip\
\newblock
\APACrefYearMonthDay{1976}{}{}.
\newblock
{\BBOQ}\APACrefatitle {Distance measures for speech recognition, psychological
  and instrumental} {Distance measures for speech recognition, psychological
  and instrumental}.{\BBCQ}
\newblock
\APACjournalVolNumPages{Pattern recognition and artificial
  intelligence}{116}{}{374--388}.
\PrintBackRefs{\CurrentBib}

\bibitem [\protect \citeauthoryear {%
Orena%
, Theodore%
\BCBL {}\ \BBA {} Polka%
}{%
Orena%
\ \protect \BOthers {.}}{%
{\protect \APACyear {2015}}%
}]{%
orena2015language}
\APACinsertmetastar {%
orena2015language}%
\begin{APACrefauthors}%
Orena, A\BPBI J.%
, Theodore, R\BPBI M.%
\BCBL {}\ \BBA {} Polka, L.%
\end{APACrefauthors}%
\unskip\
\newblock
\APACrefYearMonthDay{2015}{}{}.
\newblock
{\BBOQ}\APACrefatitle {Language exposure facilitates talker learning prior to
  language comprehension, even in adults} {Language exposure facilitates talker
  learning prior to language comprehension, even in adults}.{\BBCQ}
\newblock
\APACjournalVolNumPages{Cognition}{143}{}{36--40}.
\PrintBackRefs{\CurrentBib}

\bibitem [\protect \citeauthoryear {%
Perea%
\ \protect \BOthers {.}}{%
Perea%
\ \protect \BOthers {.}}{%
{\protect \APACyear {2014}}%
}]{%
perea2014ability}
\APACinsertmetastar {%
perea2014ability}%
\begin{APACrefauthors}%
Perea, M.%
, Jim{\'e}nez, M.%
, Su{\'a}rez-Coalla, P.%
, Fern{\'a}ndez, N.%
, Vi{\~n}a, C.%
\BCBL {}\ \BBA {} Cuetos, F.%
\end{APACrefauthors}%
\unskip\
\newblock
\APACrefYearMonthDay{2014}{}{}.
\newblock
{\BBOQ}\APACrefatitle {Ability for voice recognition is a marker for dyslexia
  in children} {Ability for voice recognition is a marker for dyslexia in
  children}.{\BBCQ}
\newblock
\APACjournalVolNumPages{Experimental Psychology}{}{}{}.
\PrintBackRefs{\CurrentBib}

\bibitem [\protect \citeauthoryear {%
T.~Perrachione%
, Dougherty%
, McLaughlin%
\BCBL {}\ \BBA {} Lember%
}{%
T.~Perrachione%
\ \protect \BOthers {.}}{%
{\protect \APACyear {2015}}%
}]{%
perrachione2015effects}
\APACinsertmetastar {%
perrachione2015effects}%
\begin{APACrefauthors}%
Perrachione, T.%
, Dougherty, S.%
, McLaughlin, D.%
\BCBL {}\ \BBA {} Lember, R.%
\end{APACrefauthors}%
\unskip\
\newblock
\APACrefYearMonthDay{2015}{}{}.
\newblock
{\BBOQ}\APACrefatitle {The effects of speech perception and speech
  comprehension on talker identification.} {The effects of speech perception
  and speech comprehension on talker identification.}{\BBCQ}
\newblock
\BIn{} \APACrefbtitle {ICPhS.} {Icphs.}
\PrintBackRefs{\CurrentBib}

\bibitem [\protect \citeauthoryear {%
T\BPBI K.~Perrachione%
}{%
T\BPBI K.~Perrachione%
}{%
{\protect \APACyear {2018}}%
}]{%
perrachione_2018}
\APACinsertmetastar {%
perrachione_2018}%
\begin{APACrefauthors}%
Perrachione, T\BPBI K.%
\end{APACrefauthors}%
\unskip\
\newblock
\APACrefYearMonthDay{2018}{{\APACmonth{12}}}{}.
\newblock
{\BBOQ}\APACrefatitle {Recognizing {Speakers} {Across} {Languages}}
  {Recognizing {Speakers} {Across} {Languages}}.{\BBCQ}
\newblock
\BIn{} S.~Frühholz\ \BBA {} P.~Belin\ (\BEDS), \APACrefbtitle {The {Oxford}
  {Handbook} of {Voice} {Perception}} {The {Oxford} {Handbook} of {Voice}
  {Perception}}\ (\BPGS\ 514--538).
\newblock
\APACaddressPublisher{}{Oxford University Press}.
\newblock
\begin{APACrefDOI} \doi{10.1093/oxfordhb/9780198743187.013.23} \end{APACrefDOI}
\PrintBackRefs{\CurrentBib}

\bibitem [\protect \citeauthoryear {%
Povey%
\ \protect \BOthers {.}}{%
Povey%
\ \protect \BOthers {.}}{%
{\protect \APACyear {2011}}%
}]{%
povey2011kaldi}
\APACinsertmetastar {%
povey2011kaldi}%
\begin{APACrefauthors}%
Povey, D.%
, Ghoshal, A.%
, Boulianne, G.%
, Burget, L.%
, Glembek, O.%
, Goel, N.%
\BDBL {}others%
\end{APACrefauthors}%
\unskip\
\newblock
\APACrefYearMonthDay{2011}{}{}.
\newblock
{\BBOQ}\APACrefatitle {The Kaldi speech recognition toolkit} {The kaldi speech
  recognition toolkit}.{\BBCQ}
\newblock
\BIn{} \APACrefbtitle {IEEE 2011 workshop on automatic speech recognition and
  understanding.} {Ieee 2011 workshop on automatic speech recognition and
  understanding.}
\PrintBackRefs{\CurrentBib}

\bibitem [\protect \citeauthoryear {%
Schatz%
\ \protect \BOthers {.}}{%
Schatz%
\ \protect \BOthers {.}}{%
{\protect \APACyear {2013}}%
}]{%
schatz2013evaluating}
\APACinsertmetastar {%
schatz2013evaluating}%
\begin{APACrefauthors}%
Schatz, T.%
, Peddinti, V.%
, Bach, F.%
, Jansen, A.%
, Hermansky, H.%
\BCBL {}\ \BBA {} Dupoux, E.%
\end{APACrefauthors}%
\unskip\
\newblock
\APACrefYearMonthDay{2013}{}{}.
\newblock
{\BBOQ}\APACrefatitle {Evaluating speech features with the minimal-pair ABX
  task: Analysis of the classical MFC/PLP pipeline} {Evaluating speech features
  with the minimal-pair abx task: Analysis of the classical mfc/plp
  pipeline}.{\BBCQ}.
\PrintBackRefs{\CurrentBib}

\bibitem [\protect \citeauthoryear {%
Thompson%
}{%
Thompson%
}{%
{\protect \APACyear {1987}}%
}]{%
thompson1987language}
\APACinsertmetastar {%
thompson1987language}%
\begin{APACrefauthors}%
Thompson, C\BPBI P.%
\end{APACrefauthors}%
\unskip\
\newblock
\APACrefYearMonthDay{1987}{}{}.
\newblock
{\BBOQ}\APACrefatitle {A language effect in voice identification} {A language
  effect in voice identification}.{\BBCQ}
\newblock
\APACjournalVolNumPages{Applied Cognitive Psychology}{1}{2}{121--131}.
\PrintBackRefs{\CurrentBib}

\bibitem [\protect \citeauthoryear {%
Thorburn%
, Feldman%
\BCBL {}\ \BBA {} Schatz%
}{%
Thorburn%
\ \protect \BOthers {.}}{%
{\protect \APACyear {2019}}%
}]{%
thorburn_quantitative_2019}
\APACinsertmetastar {%
thorburn_quantitative_2019}%
\begin{APACrefauthors}%
Thorburn, C\BPBI A.%
, Feldman, N\BPBI H.%
\BCBL {}\ \BBA {} Schatz, T.%
\end{APACrefauthors}%
\unskip\
\newblock
\APACrefYearMonthDay{2019}{}{}.
\newblock
{\BBOQ}\APACrefatitle {A quantitative model of the language familiarity effect
  in infancy} {A quantitative model of the language familiarity effect in
  infancy}.{\BBCQ}
\newblock
\BIn{} \APACrefbtitle {Proceedings of the Conference on Cognitive Computational
  Neuroscience.} {Proceedings of the conference on cognitive computational
  neuroscience.}
\PrintBackRefs{\CurrentBib}

\bibitem [\protect \citeauthoryear {%
Wester%
}{%
Wester%
}{%
{\protect \APACyear {2010}}%
}]{%
wester2010emime}
\APACinsertmetastar {%
wester2010emime}%
\begin{APACrefauthors}%
Wester, M.%
\end{APACrefauthors}%
\unskip\
\newblock
\APACrefYearMonthDay{2010}{}{}.
\newblock
\APACrefbtitle {The EMIME bilingual database} {The emime bilingual database}\
  \APACbVolEdTR{}{\BTR{}}.
\newblock
\APACaddressInstitution{}{The University of Edinburgh}.
\PrintBackRefs{\CurrentBib}

\bibitem [\protect \citeauthoryear {%
Wester%
}{%
Wester%
}{%
{\protect \APACyear {2012}}%
}]{%
wester2012talker}
\APACinsertmetastar {%
wester2012talker}%
\begin{APACrefauthors}%
Wester, M.%
\end{APACrefauthors}%
\unskip\
\newblock
\APACrefYearMonthDay{2012}{}{}.
\newblock
{\BBOQ}\APACrefatitle {Talker discrimination across languages} {Talker
  discrimination across languages}.{\BBCQ}
\newblock
\APACjournalVolNumPages{Speech Communication}{54}{6}{781--790}.
\PrintBackRefs{\CurrentBib}

\bibitem [\protect \citeauthoryear {%
Zarate%
, Tian%
, Woods%
\BCBL {}\ \BBA {} Poeppel%
}{%
Zarate%
\ \protect \BOthers {.}}{%
{\protect \APACyear {2015}}%
}]{%
zarate2015multiple}
\APACinsertmetastar {%
zarate2015multiple}%
\begin{APACrefauthors}%
Zarate, J\BPBI M.%
, Tian, X.%
, Woods, K\BPBI J.%
\BCBL {}\ \BBA {} Poeppel, D.%
\end{APACrefauthors}%
\unskip\
\newblock
\APACrefYearMonthDay{2015}{}{}.
\newblock
{\BBOQ}\APACrefatitle {Multiple levels of linguistic and paralinguistic
  features contribute to voice recognition} {Multiple levels of linguistic and
  paralinguistic features contribute to voice recognition}.{\BBCQ}
\newblock
\APACjournalVolNumPages{Scientific reports}{5}{1}{1--9}.
\PrintBackRefs{\CurrentBib}

\end{thebibliography}

\end{document}